# Personal Protective Equipment Detection in Extreme Construction Conditions

**Yuexiong Ding;[1] Xiaowei Luo, M.ASCE[2]**

[1]Department of Architecture and Civil Engineering, City University of Hong Kong, Hong Kong, China. Email: yxding7-c@my.cityu.edu.hk
[2]Department of Architecture and Civil Engineering, City University of Hong Kong, Hong Kong, China. Email: xiaowluo@cityu.edu.hk

## ABSTRACT

Object detection has been widely applied for construction safety management, especially personal protective equipment (PPE) detection. Though the existing PPE detection models trained on conventional datasets have achieved excellent results, their performance dramatically declines in extreme construction conditions. A robust detection model NST-YOLOv5 is developed by combining the neural style transfer (NST) and YOLOv5 technologies. Five extreme conditions are considered and simulated via the NST module to endow the detection model with excellent robustness, including low light, intense light, sand dust, fog, and rain. Experiments show that the NST has great potential as a tool for extreme data synthesis since it is better at simulating extreme conditions than other traditional image processing algorithms and helps the NST-YOLOv5 achieve 0.141 and 0.083 $mAP_{05:95}$ improvements in synthesized and real-world extreme data. This study provides a new feasible way to obtain a more robust detection model for extreme construction conditions.

## INTRODUCTION

Most construction casualties can be avoided if workers wear appropriate personal protective equipment (PPE). Therefore, one of the on-site safety inspection tasks is to check whether the worker is wearing appropriate PPE. Many artificial intelligence (AI) applications for automated PPE detection have also been developed and applied to construction projects to assist safety officers in checking workers' PPE compliance in real-time (Nath et al. 2020; Wang et al. 2021). However, training data used in these existing PPE detection models were taken in an environment with conventional conditions, making those models lack good robustness to extreme conditions. In fact, the site environment is generally complicated, changeable, and uncontrollable, resulting in extreme working conditions from time to time. Therefore, the performance of those PPE detection models declines sharply and gives unsatisfactory detection results when extreme site conditions occur. Besides, the lack of related images in extreme conditions is another crucial factor that hinders the improvement of the current situation.

To this end, a more robust PPE detection model is developed by combining the neural style transfer (NST) and YOLOv5 technologies. Five extreme conditions are considered to verify the robustness of the final trained detection model: low light, intense light, sand dust, fog, and rain. The novelties and significance of this study can be summarized as follows: 1) exposes the dramatic performance decline of the model trained on conventional data when encountering extreme conditions; 2) develops a more robust model for PPE detection in extreme construction conditions by a new feasible way; 3) proves the superiority of using NST for extreme conditions simulation. The rest of the paper is organized as follows: section 2 reviews related works, and section 3 describes the proposed method for robust model development. Section 4 conducts data collection and experiments, as well as result analysis. Finally, section 5 concludes the study.

## LITERATURE REVIEW

With the development of deep learning, many object detection models based on the convolutional neural network (CNN) have been widely used for PPE detection, including fast region-based CNN (Fast R-CNN) (Mohammad et al. 2020), single-shot detector (SSD) (Wu et al. 2019), and you only look once (YOLO) (Nath et al. 2020; Wang et al. 2021). For example, Wu et al. (2019) applied the SSD to detect helmets and the corresponding colors, while Mohammad et al. (2020) adopted two Fast R-CNN models to detect human bodies and PPEs. Wang et al. (2021) showed that the YOLOv5 achieved state-of-the-art (SOTA) performance in accuracy and speed compared with the Fast R-CNN and SSD. However, these PPE detection models are all trained on conventional datasets, resulting in low robustness for adverse conditions. On the other hand, there is no relevant extreme data for robust PPE detection modeling, hindering the improvement of the current situation.

Deep learning-based object detection models are typically data-driven. The more images a model has seen, the more accurate the model is. Therefore, one of the best ways to improve the robustness of the detection model to extreme conditions is to collect or simulate as many related images as possible to train the model. For example, Kang et al. (2022) applied five traditional image processing algorithms (TAs) to simulate five weather conditions to develop a reliable monitoring system for construction objects. However, the effect of TAs is still limited since they only help the detection model achieve a 0.05 $mAP_{05:95}$ improvement. In addition to TAs, the NST is another excellent method for extreme conditions simulation, which achieves extreme scene embedding according to the input style and content images (Ghiasi et al. 2017). The arbitrary NST allows the trained model to transfer any given style into any content image. Benefiting from the arbitrary attribute, a significant number of extreme data can be synthesized based on a small number of conventional content and style images. The NST has recently been applied in many classification research as a data augmentation method (Jackson et al. 2019; Zheng et al. 2019), but it has not yet been explored for extreme condition simulations.

## METHODOLOGY

Figure 1 shows the overall research workflows: 1) data acquisition, 2) robust PPE detection modeling, and 3) extreme test. First, three types of images are collected: style images related to five extreme conditions, conventional images for PPE detection (PPE images), and real-world images with extreme conditions. Then, an NST-YOLOv5 is trained on a series of synthesized images simulating extreme conditions to complete robust PPE detection modeling. Finally, extreme tests are carried out using synthesized and real-world images to evaluate the robustness of the NST-YOLOv5.

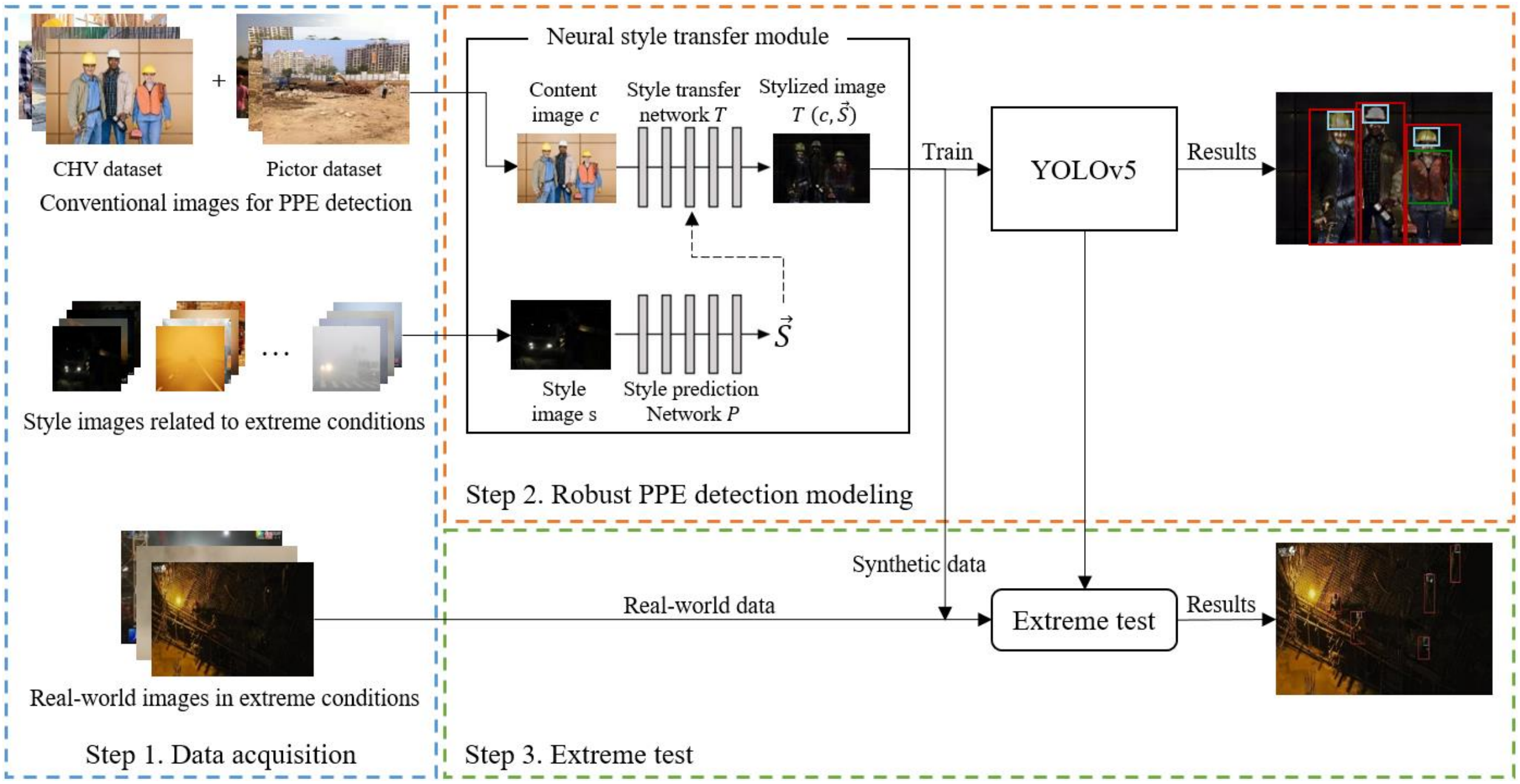

**Figure 1. Overview of the research workflows**

**Extreme condition simulation.** The NST is a superior technology for extreme conditions simulation since it transfers only the extreme style from the style image to the content image with minimized content changes. The arbitrary NST model with modular structure (Ghiasi et al. 2017) was adopted in this study, which divides the task into style extraction and style transfer and assigns them to two different networks for execution, as shown in Figure 1. Sequential execution of these two networks to complete image stylization: extract style ($\vec{S}$) from the style image via the style prediction network ($P$), then stylize the content image using the style transfer network ($T$) under the style constraints ($\vec{S}$). Such modular structure allows not only arbitrary stylization but also easy control of the stylization strength by setting the strength factor ($\alpha$) on the vector $\vec{S}$ when executing network $T$. Theoretically, the total number of stylized extreme images ($N_E$) is equal to $N_c \times N_s \times N_\alpha$, where $N_c$, $N_s$, and $N_\alpha$ are the number of content images, extreme-style images, and stylization strength factors, respectively. Therefore, the $N_E$ can be expanded by increasing the $N_c$ and $N_s$, alleviating the problem of lacking extreme data for robust PPE detection modeling since conventional PPE images and extreme-style images are much more accessible than extreme PPE images.

**Robust PPE detection modeling.** As shown in step 2 of Figure 1, the robust PPE detection modeling in this study is realized by combining technologies of the NST and YOLOv5, named NST-YOLOv5. As reviewed in section 2, the YOLO is currently one of the most common and popular networks for object detection, and the YOLOv5 has achieved SOTA performance in construction-related studies (Wang et al. 2021). The YOLOv5 has several improvements over the previous version YOLOv4 (Jocher et al. 2021). First, YOLOv5 ad a new Focus module to reduce parameters and increase forward and backward calculation speed. Second, the Cross Stage Partial (CSP) technique is utilized in YOLOv5's Neck structure, improving the feature extraction ability in the Neck stage compared with the ordinary stacking of convolution modules in YOLOv4. Finally, YOLOv5 adopts K-means and genetic learning algorithms to automatically learn the bounding box anchors for different datasets, which is quite helpful in making sure the most appropriate anchor boxes are used for the specific dataset to have better detection results. In this study, the NST and YOLOv5 were launched simultaneously as a whole network in the training and validation stages, while only the YOLOv5 was set to trainable. During each training or validation iteration, three random values were generated to help the NST module determine the desired extreme style type, style image, and stylization strength for each batch of conventional image input. However, the NST module will not be loaded once the NST-YOLOv5 finishes training, keeping the NST-YOLOv5 at the same inference speed as the stander YOLOv5.

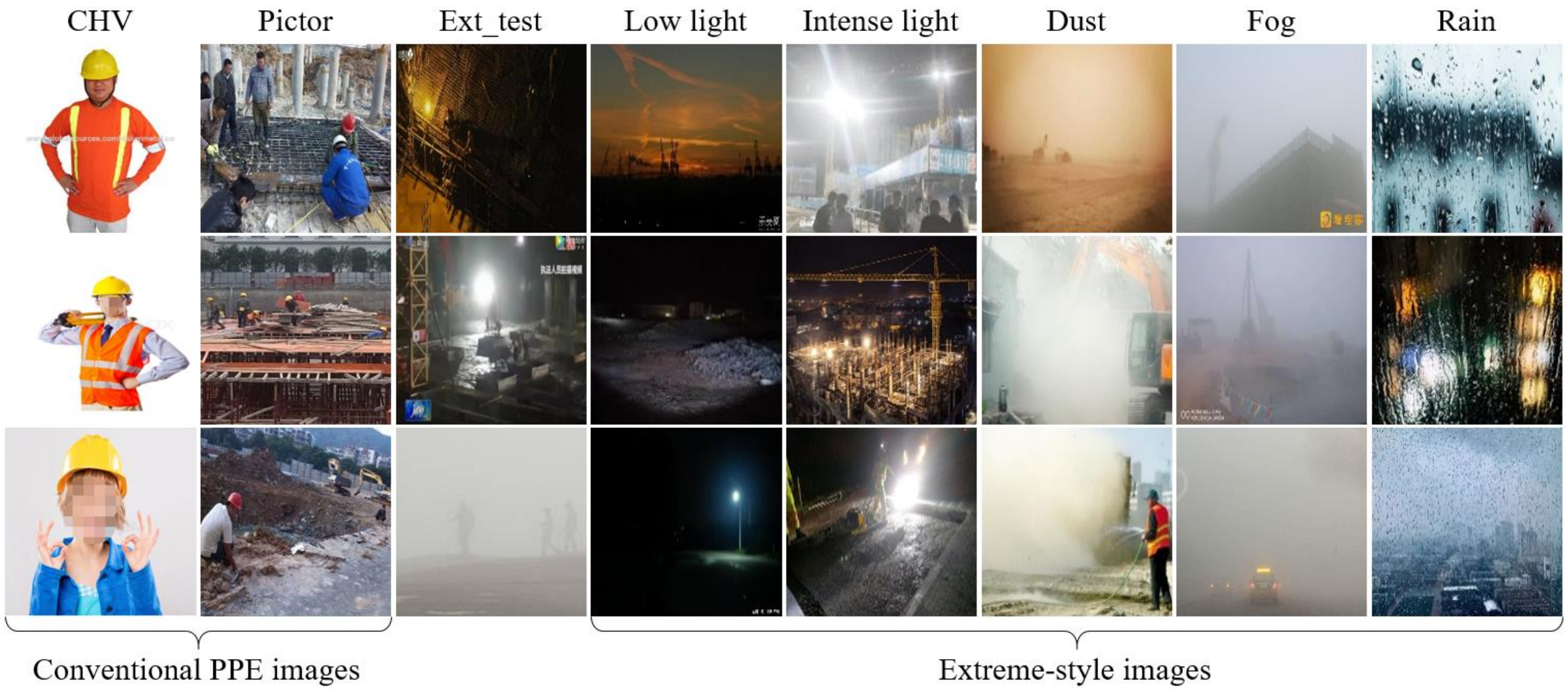

**Figure 2. Some sample images of each dataset.**

## EXPERIMENTS

**Data acquisition.** Two public datasets annotated with three categories (worker, vest, and helmet), CHV (Wang et al. 2021) and Pictor (Nath et al. 2020), were adopted in this study to form the conventional dataset (CHV_Pictor) for PPE detection modeling. Following and combining the data split results of the CHV and Pictor datasets, the CHV_Pictor dataset had 1566, 253, and 285 images for training, validation, and testing, respectively. The extreme-style image set contained a total of 100 images related to five considered extreme conditions: low light, intense light, sand dust, fog, and rain, which was collected from the Internet of 20 images per condition. Finally, 101 real-world extreme PPE images (Ext_test) covering the five considered extreme conditions were additionally collected from the Internet and annotated manually to conduct the extreme test. Figure 2 shows some sample images of each dataset used in this study.

**Results and analysis.** The implementation details of the NST-YOLOv5 modeling are as follows: the image size was $640 \times 640$, the batch size was 4, and the optimizer was SGD with a learning rate of 0.01. The YOLOv5x was adopted as the detection model (still called YOLOv5 below). Two MS COCO standard metrics of average precision (AP), $AP_{05}$ and $AP_{05:95}$, were applied for evaluation. All experiments were conducted in the "Python 3.6 + RTX 2080 Ti" environment.

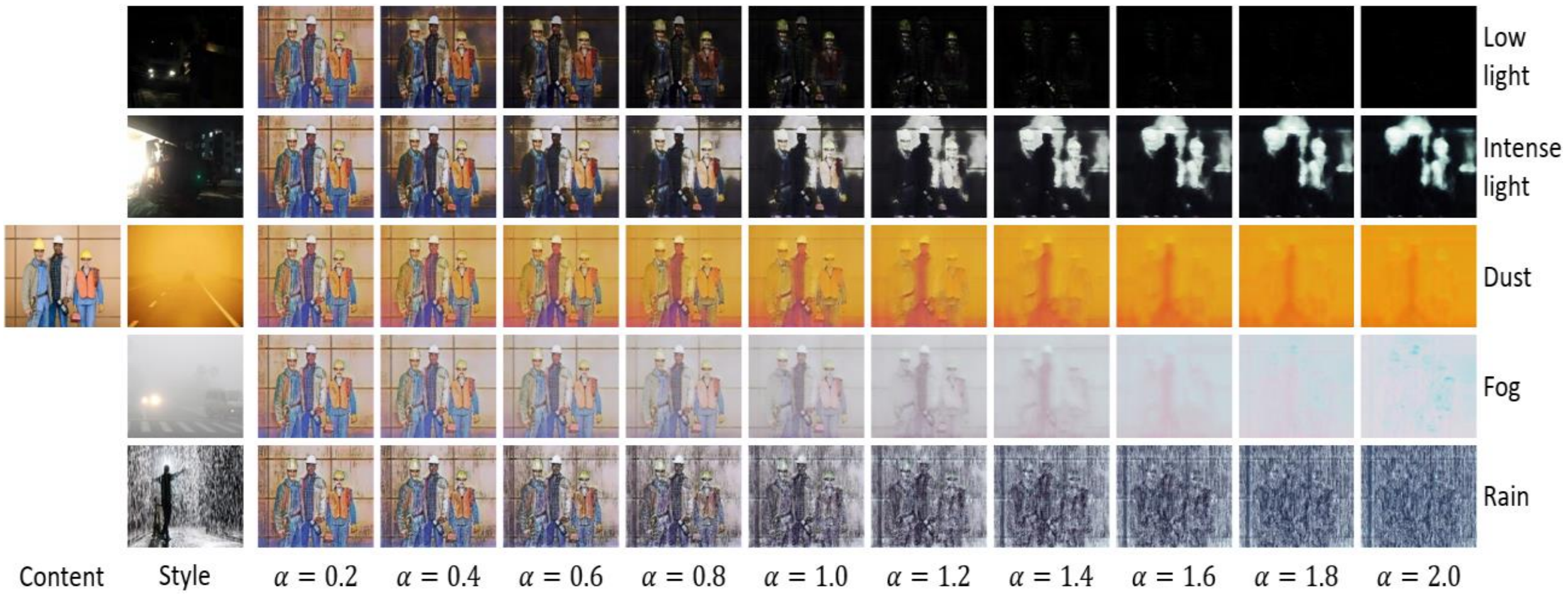

**Figure 3. Extreme condition simulation using the NST in different stylization strengths ($\alpha$).**

<u>*Stylization strength determination.*</u> A well-trained stylization model developed by Google Brain Team (Magenta Team 2017) was used as the NST module of the NST-YOLOv5. Figure 3 shows some stylized images using different extreme styles and stylization strengths ($\alpha$). The stylized results are almost identical to the original content images when $\alpha < 1.0$. As the $\alpha$ continues to increase, the content of the original image becomes increasingly difficult to be recognized by human eyes, making higher-quality simulations of extreme real-world conditions. Though the suggestion range of the $\alpha$ is [0, 1.0], the practical results show better simulations of extreme conditions when $\alpha > 1.0$ based on the human-eye observation. Therefore, the optional $\alpha$ value set was set as [0, 1.0, 1.2, 1.4, 1.6, 1.8, 2.0] when training the NST-YOLOv5, where $\alpha = 0$ means no stylization.

**Table 1. Per-class AP of the baseline and NST-YOLOv5 model on three test datasets.**

| Test set | Metric | Baseline YOLOv5 | | | | NST-YOLOv5 | | | |
|---|---|---|---|---|---|---|---|---|---|
| | | Worker | Vest | Helmet | **mAP** | Worker | Vest | Helmet | **mAP** |
| CPT | $AP_{05}$ | 0.840 | 0.812 | 0.843 | 0.832 | 0.867 | 0.854 | 0.872 | 0.864 |
| | $AP_{05:95}$ | 0.525 | 0.475 | 0.473 | 0.491 | 0.571 | 0.512 | 0.500 | 0.528 |
| SCPT | $AP_{05}$ | 0.528 | 0.483 | 0.513 | 0.508 | 0.745 | 0.724 | 0.701 | 0.723 |
| | $AP_{05:95}$ | 0.297 | 0.270 | 0.268 | 0.278 | 0.456 | 0.420 | 0.381 | 0.419 |
| Ext_test | $AP_{05}$ | 0.586 | 0.388 | 0.534 | 0.502 | 0.758 | 0.432 | 0.648 | 0.612 |
| | $AP_{05:95}$ | 0.337 | 0.177 | 0.240 | 0.251 | 0.454 | 0.236 | 0.312 | 0.334 |

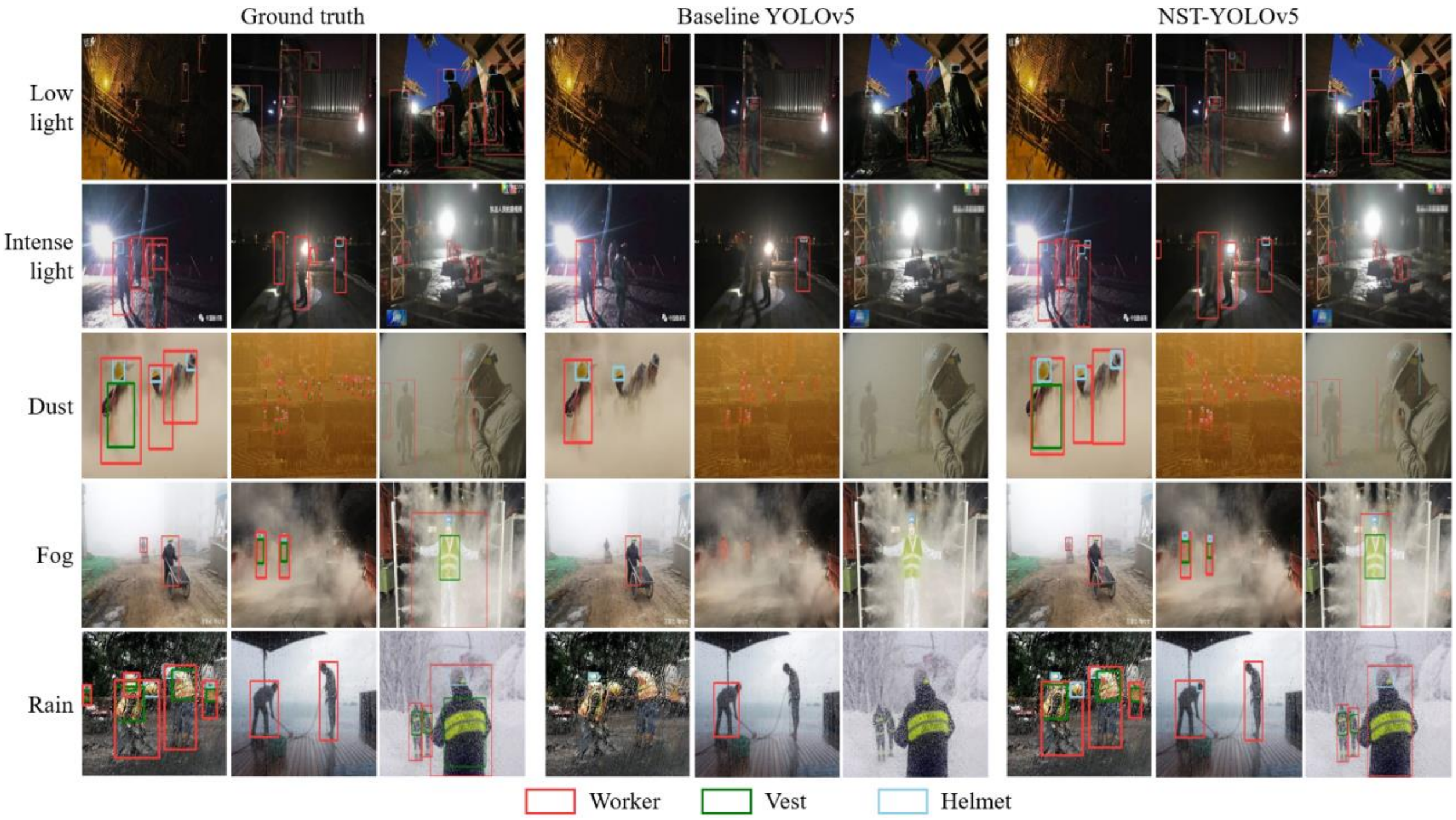

**Figure 4. PPE detection results of the NST-YOLOv5 on the Ext_test set.**

<u>*PPE detection results.*</u> Table 1 shows the PPE detection performance of the baseline YOLOv5 and NST-YOLOv5 models on three test datasets. The baseline YOLOv5 refers to a stander YOLOv5x trained on the conventional CHV_Pictor dataset. The CPT dataset denotes the CHV_Pictor_test set, and the SCPT is the stylized CPT using $\alpha = [1.0, 1.4, 1.8]$. According to results in Table 1, both the baseline YOLOv5 and NST-YOLOv5 achieve encouraging performance on conventional data (CPT) with the $mAP_{05}$ of {0.832, 0.864} and $mAP_{05:95}$ of {0.491, 0.528}. However, the performance of the baseline YOLOv5 drops sharply on the SCPT and Ext_test with $mAP_{05}$ declines of {0.324, 0.33} and $mAP_{05:95}$ declines of {0.213, 0.24}, indicating the robustness of the model trained on the conventional dataset is too poor to work well in extreme conditions. The situation is alleviated in the NST-YOLOv5 model, which improves the PPE detection performance not only on the conventional test set (CPT) but also on the other two extreme test sets (SCPT and Ext_test), with $mAP_{05}$ improvements of {0.032, 0.215, 0.11} and $mAP_{05:95}$ improvements of {0.037, 0.141, 0.083}. Some examples of PPE detection results on the Ext_test set were visualized to highlight the robustness of the NST-YOLOv5 to extreme conditions more intuitively, as shown in Figure 4. The baseline YOLOv5 almost loses the detection ability in extreme conditions, which only identifies a few prominent objects and misses most others. Promisingly, the NST-YOLOv5 detects almost all objects of interest regardless of extreme conditions, achieving a very close level to the human.

*Ablation study.* Some ablation experiments were further conducted to show the impact of applying different simulation approaches and detection backbones. The simulation method based on TAs (Buslaev et al. 2020) and the other two other well-known object detection models, Faster R-CNN and SSD, were adopted in this section for comparison study. The ablation results are shown in Table 2. First, YOLOv5-based models perform best on all test sets compared with other types of object detection models. Second, the Faster R-CNN and SSD models also have sharp performance declines on the extreme datasets (SCPT and Ext_Test), highlighting the universality of the vulnerability to extreme conditions of the model trained only on conventional data. Third, the NST-YOLOv5 outperforms the TA-YOLOv5, indicating that the NST is better at simulating extreme conditions than other traditional synthesis methods. Finally, the NST also works efficiently for the Faster R-CNN and SSD, showing its generalization as a potential tool for extreme data synthesis and augmentation.

**Table 2. Performance comparison of different models on three test datasets.**

| Models | CPT | | SCPT | | Ext_Test | |
|---|---|---|---|---|---|---|
| | $AP_{05}$ | $AP_{05:95}$ | $AP_{05}$ | $AP_{05:95}$ | $AP_{05}$ | $AP_{05:95}$ |
| Faster R-CNN | 0.624 | 0.323 | 0.205 | 0.09 | 0.196 | 0.075 |
| SSD | 0.779 | 0.406 | 0.291 | 0.14 | 0.309 | 0.132 |
| YOLOv5 | 0.832 | 0.491 | 0.508 | 0.278 | 0.502 | 0.251 |
| TA-YOLOv5 | 0.849 | 0.516 | 0.594 | 0.287 | 0.54 | 0.254 |
| NST-Faster R-CNN | 0.664 | 0.350 | 0.476 | 0.237 | 0.256 | 0.108 |
| NST-SSD | 0.780 | 0.409 | 0.598 | 0.298 | 0.411 | 0.185 |
| NST-YOLOv5 | 0.864 | 0.528 | 0.723 | 0.419 | 0.612 | 0.334 |

## CONCLUSION

This study develops an NST-YOLOv5 model for robust PPE detection of construction workers. The combination of NST to the YOLOv5 successfully transfers the issue of extreme data shortage to the acquisition of conventional PPE and extreme-style images. Comparison experiments between the baseline YOLOv5 and NST-YOLOv5 models show that the model trained only on conventional data is vulnerable to extreme conditions, while the NST-YOLOv5 achieved significant robustness improvement with 0.141 and 0.083 $mAP_{05:95}$ increases on the synthesized and real-world extreme images. The ablation experiments further indicate that detection models' vulnerability to extreme conditions is universal when training only on conventional data, while the NST module is a better general tool to endow various detection models with great robustness. Future work can further explore the optimal $\alpha$ value set for extreme conditions simulations.

Besides, developing a detection model with image adaptation is also desirable, which adaptively eliminates extreme condition information from the image to achieve an optimal image state for the subsequent object detection.

## ACKNOWLEDGEMENT

The Shenzhen Science and Technology Innovation Committee Grant #JCYJ20180507181647320 and General Research Fund from Research Grant Council of Hong Kong SAR # 11211622 jointly supported this work. The conclusions herein are those of the authors and do not necessarily reflect the views of the sponsoring agencies.